\documentclass[IEEEtran, 10pt]{article}
\usepackage{amsmath} 
\usepackage{amsthm} 
\usepackage{amsfonts} 
\usepackage{amssymb}
\usepackage{multirow}
\usepackage{dcolumn}
\usepackage{array}
\usepackage{graphicx}
\usepackage{titlesec}
\usepackage{enumitem}
\usepackage{url}
\usepackage{authblk}
\usepackage{graphicx}
\usepackage{caption}
\usepackage{booktabs}
\usepackage{verbatim}

\usepackage{tcolorbox}

\usepackage{xcolor}
\usepackage{soul}

\usepackage{tikz}
\usepackage{pgfplots}
\usepgfplotslibrary{units}

\usepackage{cite} 

\titleformat*{\subsubsection}{\large }
\titleformat*{\paragraph}{\normalsize \itshape}

\usepackage[T1]{fontenc}
\usepackage{txfonts}
\usepackage{fixltx2e}
\usepackage{tabu}
\newcolumntype{C}[1]{>{\centering\let\newline\\\arraybackslash\hspace{0pt}}m{#1}}

\usepackage[margin=1in]{geometry}
\usepackage{color,hyperref}
\definecolor{darkblue}{rgb}{0.0,0.0,0.3}
\hypersetup{colorlinks,breaklinks,
linkcolor=darkblue,urlcolor=darkblue,
anchorcolor=darkblue,citecolor=darkblue}
\usepackage{url}
\usepackage{graphicx}

\date{}
\newlist{myenumi}{description}{10}
\setlist[myenumi]{labelindent=\parindent, leftmargin=*, label=(\arabic*), align=left}
\setlist[myenumi]{leftmargin=15pt}

\usepackage{setspace}
\begin{document}
%\pagestyle{fancy}
%\begin{center}
%{ Scooner}\\
%\vspace*{0.3cm}
%\end{center}
%\vspace*{0.3cm}
%\input{title_page}
\title{\Large Comparison of pipeline, sequence-to-sequence, and GPT models for end-to-end relation extraction: experiments with the rare disease use-case}
\author[2]{Shashank Gupta}
\author[2]{Xuguang Ai}
\author[1, 2]{Ramakanth Kavuluru}
\affil[1]{Division of Biomedical Informatics, Dept.~of Internal Medicine, University of Kentucky, USA}
\affil[2]{Computer Science Department, University of Kentucky, USA}
%\affil[3]{IBM Research Australia, Melbourne, Australia}
%\titlelabel{\large \thesection \quad}
\maketitle
\thispagestyle{empty}

\begin{abstract}
\noindent{\bf Objective: } End-to-end relation extraction (E2ERE) is an important and realistic application of natural language processing (NLP) in biomedicine. In this paper, 
we aim to compare three prevailing paradigms for E2ERE using a complex dataset focused on rare diseases involving discontinuous and nested entities. 

\noindent{\bf  Methods:}  We use the \textbf{\texttt{RareDis}} information extraction dataset to evaluate three competing approaches (for E2ERE): NER $\rightarrow$ RE pipelines,   joint sequence to sequence models, and  generative pre-trained transformer (GPT) models.   We use comparable state-of-the-art models and best practices for each of these approaches and conduct error analyses to assess their failure modes.

\noindent{\bf Results: }  Our findings reveal that pipeline models are still the best, while sequence-to-sequence models are not far behind; GPT models with eight times as many parameters are worse than even sequence-to-sequence models and lose to pipeline models by over 10 F1 points. Partial matches and discontinuous entities caused many NER errors contributing to lower overall E2E performances. We also verify these findings on a second E2ERE dataset for chemical-protein interactions. Although generative LM-based methods are more suitable for zero-shot settings, when training data is available, our results show that it is better to work with more conventional models trained and tailored for E2ERE. 

%It is not clear if GPT models that have 100 to 1000 times more parameters than pipeline models will yield better results, but the expensive hardware needed to host and fine-tune them may not be cost effective for many applications.

\noindent{\bf Conclusion: }  More innovative methods are needed to marry the best of the both worlds from smaller encoder-decoder pipeline models and the larger GPT models to improve E2ERE. As of now, we see that well designed pipeline models offer substantial performance gains at a lower cost and carbon footprint for E2ERE. 
Our contribution is also the first to conduct E2ERE for the \textbf{\texttt{RareDis}} dataset. The dataset and code for all our experiments are publicly available: \url{https://github.com/shashank140195/Raredis}

\end{abstract}

\section{INTRODUCTION}
\label{sec-intro}
Named entities and relations among them are basic units of information in many disciplines including biomedicine. A relation is typically expressed as a triple that has a subject entity and an object entity connected via a predicate (or relation type) as in the example (\texttt{subject}: atorvastatin, \texttt{predicate}: treats, \texttt{object}: hyperlipidemia). 
Disease and treatment mechanisms are often driven at the biological level by protein-protein and chemical-protein interactions while clinical relations such as drug-disease treatment relations and disease-symptom causative relations are helpful in providing care.
Most new relational information is first discussed in textual narratives (e.g., scientific literature, clinical notes, or social media posts), and extracting and storing it as triples enable effective search systems~\cite{dietze2009goweb}, high-level reasoning, hypothesis generation, and knowledge discovery applications~\cite{henry2017literature}. As such, named entity recognition (NER) and relation extraction (RE) have become standard tasks in biomedical natural language processing (BioNLP)~\cite{kilicoglu2020broad}. 

Many RE efforts in the past assume that the entity spans are already provided as part of the input and hence addressed an easier problem of relation classification (RC)~\cite{zeng2014relation,zhou2016attention,kavuluru2017extracting}. However, a more realistic setting is the ability to extract both entity spans and associated relations from the raw text where entities are not provided. RE in this setting is generally called end-to-end relation extraction (E2ERE). With the recent deluge of deep neural networks (or deep learning methods), the NLP community has been focusing more on E2ERE efforts~\cite{miwa2016end,zhang2017end,pawar2017end,tran2018end}. Efforts have also been expanded from single sentence E2ERE to a more complex setting of extractions at the document level, involving cross-sentence relations, where entities expressed in different sentences are to be linked~\cite{peng2017cross,yao2019docred}. Additional intricacies arise when named entities are discontinuous or when their spans overlap~\cite{li2021span}. For example, consider the string ``\underline{\textit{accumulation of}} fats (lipids) called \underline{\textit{GM 2 gangliosides}},'' where entity span ``accumulation of GM 2 gangliosides'' is discontinuous with a gap involving outside words. In the example phrase ``central pain syndrome,'' both the full three-word string and the middle word ``pain'' can constitute two different entities, where the latter entity is fully nested in the longer 3-word entity. Thus far, we have not seen efforts handling these complex document-level E2ERE settings involving discontinuous and overlapping/nested entities. In this paper, we address this using the recently introduced RE dataset called \textbf{\texttt{RareDis}}~\cite{martinez2022raredis}, which focuses on information extraction for rare diseases and has the complex traits indicated earlier. Although there is another dataset that focuses on rare diseases at the sentence level~\cite{fabregat2018deep}, we use \textbf{\texttt{RareDis}} since it operates at the document level.

Over the past decade, neural methods especially those involving contextual dense word embeddings have supplanted conventional NLP methods that relied on n-gram statistics. For E2ERE, joint learning neural methods that simultaneously optimized for NER and RE objectives~\cite{eberts2020span,tran2019neural} have gained popularity over pipeline-based methods that build two separate models for NER and RE, where the NER model's output is fed to the RE model. However, the recent Princeton University Relation Extraction (PURE) framework~\cite{zhong2021frustratingly} proposed an intuitive pipeline method that takes advantage of the so-called typed ``entity markers'' to encapsulate entity spans provided as input to contextualized language models (LMs). The PURE method reignited the relevance of cleverly designed pipeline methods when compared with joint learning methods. Simultaneously, sequence-to-sequence models that became popular for machine translation have been repurposed~\cite{nayak2020effective} effectively for E2ERE where the encoder-decoder architecture is used to transform raw text to directly output relations encoded through so-called ``linearization schemas'' and ``copy mechanism''~\cite{zeng2018extracting}. The state-of-the-art (SoTA) for this paradigm of models is the Seq2Rel architecture~\cite{giorgi2022sequencetosequence} that inherently allows for E2ERE. Another latest seq2seq architecture called T5 (Text-To-Text Transfer Transformer~\cite{raffel2020exploring}  and its variant Flan-T5 (Instruction finetuned version of T5)~\cite{chung2022scaling} have shown promising results in many NLP tasks needing language understanding. Finally, generative pre-trained transformers (GPTs) have gained traction and publicity (thanks to ChatGPT), especially for zero-shot and few-shot settings~\cite{radford2019language,brown2020language}. In biomedicine, BioGPT~\cite{Luo_2022} and BioMedLM~\cite{biomed-lm} have been shown to work well for relation extraction and question answering,  among generative decoder-only language models (LMs), producing SoTA scores on a few datasets. 

%\hl{Recently, the 2.7B parameter model, BioMedLM (former PubmMedGPT) by Stanford has shown SOTA scores on the medical Q\&A tasks and we repurpose it for RE.}  

Thus we identify (a) PURE for pipelines, (b) Seq2Rel, T5, and its variant Flan-T5 for sequence-to-sequence models, and (c) BioMedLM\footnote{Although we experimented with BioGPT models, they are smaller than BioMedLM and were quite inferior (more later), and as such our focus in this manuscript is more on BioMedLM, the latest and largest GPT model exclusively trained on biomedical literature}  for generative (autoregressive) LMs as representative models for prevailing competing paradigms for RE. Now the central question is, which of these approaches works well for the complex document level E2ERE task involving discontinuous and overlapping entities manifesting in the \textbf{\texttt{RareDis}} dataset? Toward answering this, we make the following contributions in this paper.
\begin{itemize}
\item We explore and provide descriptive statistics of the \textbf{\texttt{RareDis}} dataset and fix certain formatting/annotation errors in the original dataset (acknowledged by its creators) to ensure availability for the community for further benchmarking.

\item We adapt the PURE pipeline approach to the \textbf{\texttt{RareDis}} dataset since the original method does not handle discontinuous and nested entities. 

\item We design linearization schemas for the Seq2Rel method and appropriate supervised prompting strategies for T5 and BioMedLM in the context of E2ERE for the \textbf{\texttt{RareDis}} dataset. 

\item We provide quantitative evaluations of the four models (and associated variants) and conduct qualitative evaluations through manual error analyses. We make publicly available the modified \textbf{\texttt{RareDis}} dataset and code for all our experiments: \url{https://github.com/shashank140195/Raredis}
\end{itemize}

To our knowledge, our effort is the first to handle E2ERE 
with the \textbf{\texttt{RareDis}} dataset and also to compare SoTA approaches arising from three different competing paradigms in the neural RE landscape.

\begin{tcolorbox}
\begin{center} \textbf{Statement of Significance} \end{center}

\textbf{Problem}: It is not clear what NLP methods work best in practice for end-to-end relation extraction

\textbf{What is already known}: Although pipeline methods used to be the norm, recent literature shows a rise in sequence-to-sequence and decoder-only GPT models for information extraction. There is also general tendency to prefer the fancier latter models considering the excitement in the field for them.

\textbf{What this paper adds}: With the use-case of a rare disease information extraction task involving discontinuous and overlapping entities, we compare three different competing paradigms (pipeline, seq2seq, and GPT) for end-to-end relation extraction. Our findings show that a well-designed pipeline model is computationally inexpensive and more effective than other methods.

\end{tcolorbox}

\section{METHODS}
\subsection{The \textbf{\texttt{RareDis}}  dataset}
\label{sec-data}
The National Institutes of Health (NIH) estimates that around 7,000 rare diseases impact between 25 and 30 million Americans, which translates to approximately 1 out of every 10 Americans~\cite{RDD-FAQ}. Around 95\% of the known rare diseases currently lack any treatment options~\cite{RDD-FAQ}. Because these diseases are so rare, they can be challenging to diagnose and treat --- nearly 95\% of rare diseases have no known cure, and the number of drugs available for treating these conditions is limited to 100~\cite{klimova2017global}. The average diagnostic delay is around seven years~\cite{global-genes}. Many rare diseases are genetic in nature and are caused by mutations in a single gene. However, because there are thousands of rare diseases, each with unique symptoms and genetic causes, developing effective treatments can be a significant challenge. Developing a structured compendium of information about rare diseases has the potential to help expedite search, discovery, and hypothesis generation for these conditions. This necessitates developing NLP models for RE in this setting and toward this goal,  
Maritinez-deMiguel et al.~\cite{martinez2022raredis} created an annotated corpus for rare disease-related information extraction. This resource is based on the database of articles about rare diseases maintained by the National Organization for Rare Disorders (\url{https://rarediseases.org/rare-diseases/}). 
The dataset contains six entity types and six relation types and the annotation process is described in detail by the authors~\cite{martinez2022raredis}. 

\begin{figure}[h]
\centering
\includegraphics[width=0.85\linewidth]{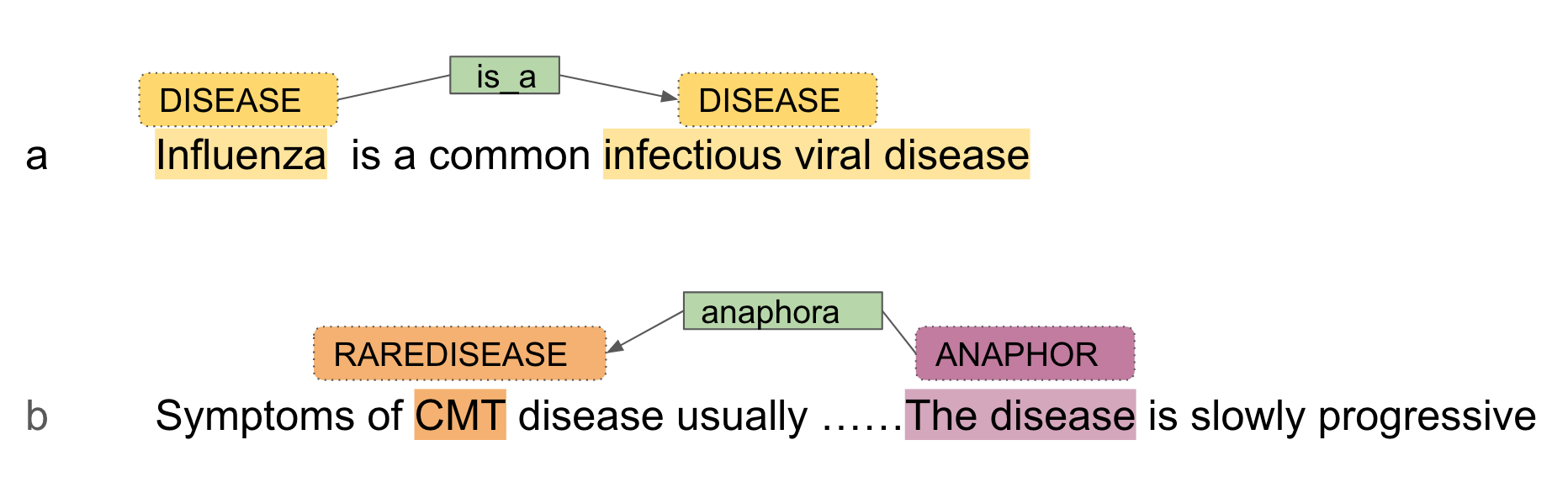}
\caption{\label{fig: ER_Example}  Examples of \textbf{is\_a} and \textbf{anaphora} relations in the \textbf{\texttt{RareDis}} dataset.}
\end{figure}

\begin{comment}
    
\begin{figure}[hb]
\centering
\includegraphics[width=\linewidth]{images/et_types_exmples_2.png}
\caption{\label{fig: et_types_example} An example of (a) Overlapped Entity (b) Nested Entity (c) Discontinous Entity in the dataset }
\end{figure}
\end{comment}

\subsubsection{Entity and relations types}
The six entity types in \textbf{\texttt{RareDis}} are: \textbf{disease}, \textbf{rare disease}, \textbf{symptom}, \textbf{sign}, \textbf{anaphor}, and \textbf{rare skin disease} with frequencies shown in the first six rows of Table~\ref{tab:ent_rel_types}. 
There are six relation types (with counts shown in the last six rows of Table~\ref{tab:ent_rel_types}): 
\textbf{produces} (relation between any disease entity and a sign/symptom produced by that entity), \textbf{increase\_risk\_of} (relation between a disease entity and another disease entity where the subject disease increases the likelihood of suffering from the object disease), \textbf{is\_a} (relation between a given disease and its classification as a more general disease), \textbf{is\_acron} (relation between an acronym and its full or expanded form), \textbf{is\_synon} (relation between two different names designating the same disease) and \textbf{anaphora} (relation of an anaphor entity with its antecedent entity). Here an anaphor entity refers to pronouns or pronominal constructs (e.g., `it'' or ``this disease'') that point to a named entity that is already mentioned in the preceding context (the ``antecedent'' of the \textbf{anaphora} relation). An example is shown in Figure~\ref{fig: ER_Example}.

\begin{table}[ht]
\centering
\renewcommand{\arraystretch}{1.2}
\begin{tabular}{l r r r r}
\toprule
\textbf{Type} & \textbf{Training} & \textbf{Dev} & \textbf{Test} \\
\midrule
sign & 2945 & 798 & 528  \\
rare disease  & 2533 & 624 & 480 \\
disease & 1369 & 278 &  230\\
anaphor & 913 & 195 & 151 \\
skin rare disease  & 393 & 58 & 45 \\
symptom & 275 & 44 & 24 \\\midrule
produces & 3256 & 850 & 556 \\
anaphora & 918 & 195 & 151 \\
is\_a & 544 & 149 & 88 \\
increase\_risk\_of & 161 & 8 & 22 \\
is\_acron & 142 & 44 & 34 \\
is\_synon & 66 & 14 & 16 \\
\bottomrule
\end{tabular}
\caption{\label{tab:ent_rel_types} Statistics of entity types (first six rows) and relation types (last six rows) in the RareDis corpus.}
\end{table}

The dataset contains discontinuous and overlapping/nested entities as discussed with examples in Section~\ref{sec-intro}; Table~\ref{tab:ent_types} throws light on the relative frequency of these situations where ``flat'' corresponds to continuous entities. 
While in both tables in this section we show training, development, and test set counts, the original dataset consisted of only training and development datasets where the authors claim to withhold the test set for a future shared task, which has not happened yet. We split up their training dataset into training and development for our experiments and their development split became our test split. 

\begin{table}[ht]
\centering
\renewcommand{\arraystretch}{1.3}
\begin{tabular}{l r r r r}
\toprule
\textbf{Dataset} & \textbf{Training} & \textbf{Dev} & \textbf{Test} \\ \midrule 
Flat & 7103 & 1666 & 1212  \\
Discontinuous & 528 & 136 & 103 \\
Overlapped & 720 & 166 &  112\\
Nested & 77 & 29 & 31 \\
\midrule
Total & 8428 & 1997 & 1458 \\
\bottomrule
\end{tabular}
\caption{\label{tab:ent_types} Counts of entity types in the corpus.}
\end{table}

\subsubsection{Modifications to the original dataset}
While exploring the dataset, we observed some annotation issues that we confirmed with the creators of the \textbf{\texttt{RareDis}} dataset through email communication. Next, we describe what they are and how we fixed them at a high level in this section. We created a custom train, validate, test split of the full dataset after fixing the following errors and made it available as a Google Drive link on our GitHub page for this project.

\begin{figure}
\centering
\includegraphics[scale=0.32]{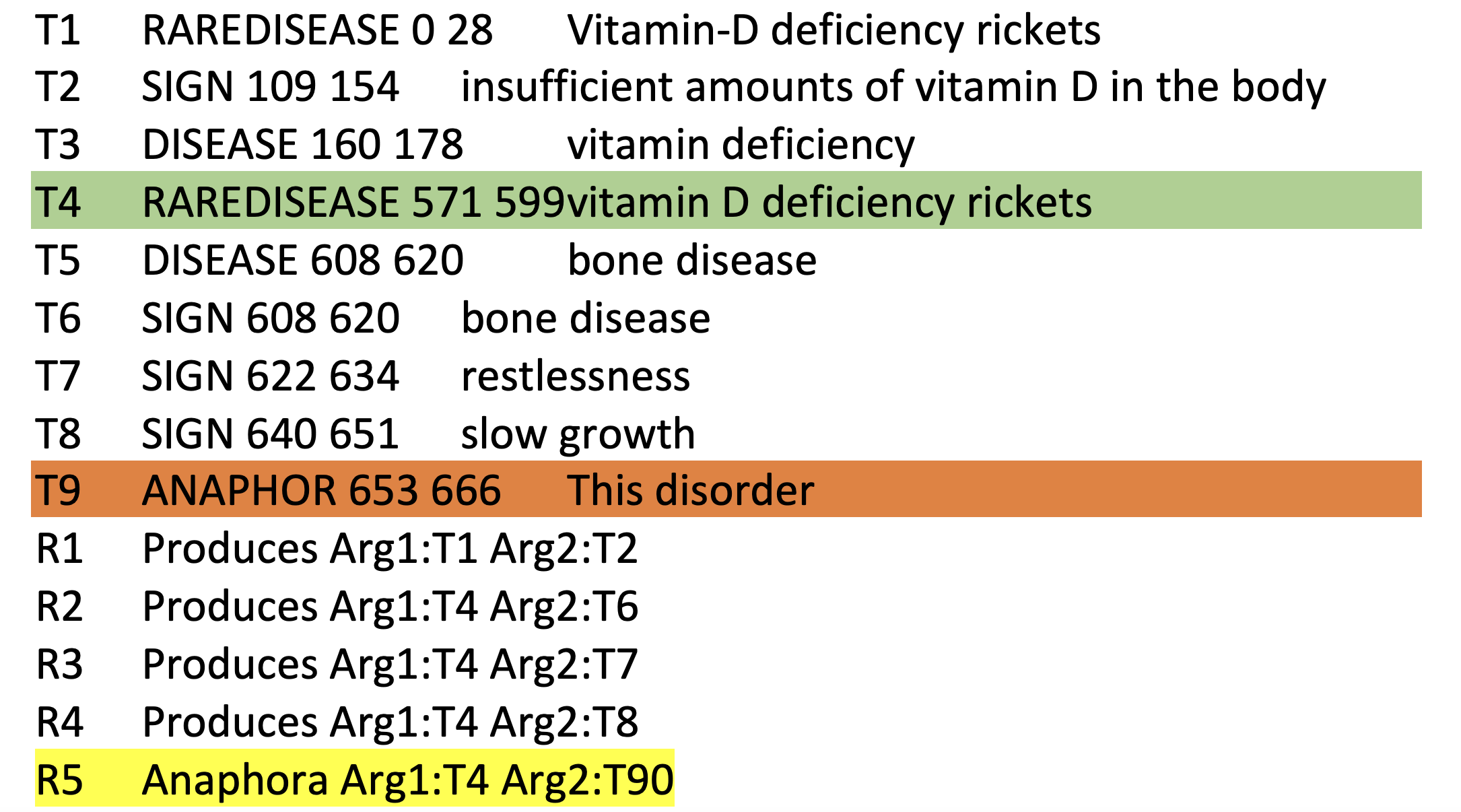}
\caption{\label{fig: RE_anno_change} An example of the argument error due to an extra trailing zero in entity IDs. Here, T90 ought to be just T9.}
\end{figure}

\paragraph{Relation argument error}
Figure~\ref{fig: RE_anno_change} shows an example of how the annotations are provided for each instance. For this example, we see the entities (T1, \ldots, T9) listed first along with types, character-based offsets, and lexical spans. Next, relations between entities are listed (R1, \ldots, R5) along with the relation type and the arguments (subject and object). Although there are only nine entities, we see for anaphora relation R5, the second argument is T90 with a trailing 0 after 9. This happened several times --- arguments in relations referring to entity IDs that are not present in the preceding entity list. This almost always happened with a trailing extra zero. We safely removed that zero and it fixed all these errors, which accounted for 9\% of the total number of relations. In the example in Figure~\ref{fig: RE_anno_change}, the anaphora relation R5 was referring to the bigram ``This disorder''.

\paragraph*{Span mismatch Error}
There were a few occasions (less than 1\% of the full dataset) where the character offsets for entities captured an extra character than needed or missed the last character of a word. We used simple rules to remove the extra character or add the missing character. For example, in the sentence ``Balantidiasis is a rare \textbf{infectious disease} caused by the single-celled (protozoan) parasite Balantidium coli,'' the bold phrase was annotated as [T24, DISEASE,1272 1289, infectious \textbf{diseas}] with a missing trailing character `e'.

\paragraph{Offset order error}
For some discontinuous entities where more than one span is part of the full entity, the order used for the spans was not left to right and we simply reordered them as such.

As outlined earlier (in Section~\ref{sec-intro}), we experiment with three different SoTA approaches each representing a competing paradigm for E2ERE. Each of these approaches is highly involved and hence we focus on high-level explanations of how they work. 

\begin{comment}
    
\begin{figure*}[ht]
\centering
\includegraphics[width=0.8\textwidth]{images/seq2rel_2.png}
\caption{\label{fig: seq2rel} A sequence-to-sequence model for document-level relation extraction. Special tokens are generated by the decoder. Entity mentions are copied from the input via a copy mechanism. Decoding is initiated by a @START@ token and terminated when the model generates the @END@ token. Attention connections are shown only for the second timestep to reduce clutter.}
\end{figure*}
\end{comment}

\subsection{The three E2ERE methods}

\subsubsection{Pipeline: The PURE Approach}
\label{sec-pipeline}
PURE by Zhong and Chen~\cite{zhong2021frustratingly} is a span-based model that has two different models for NER and RE parts of the E2ERE system. It improved upon prior joint modeling approaches even though it separately trains NER and RE models. The main argument by Zhong and Chen, the authors of PURE, is that NER and RE need different representations of tokens because they need different types of signals to make the predictions; and combining the signals can hurt the performance of both. 

\begin{figure*}[ht]
\centering
\includegraphics[width=0.98\textwidth]{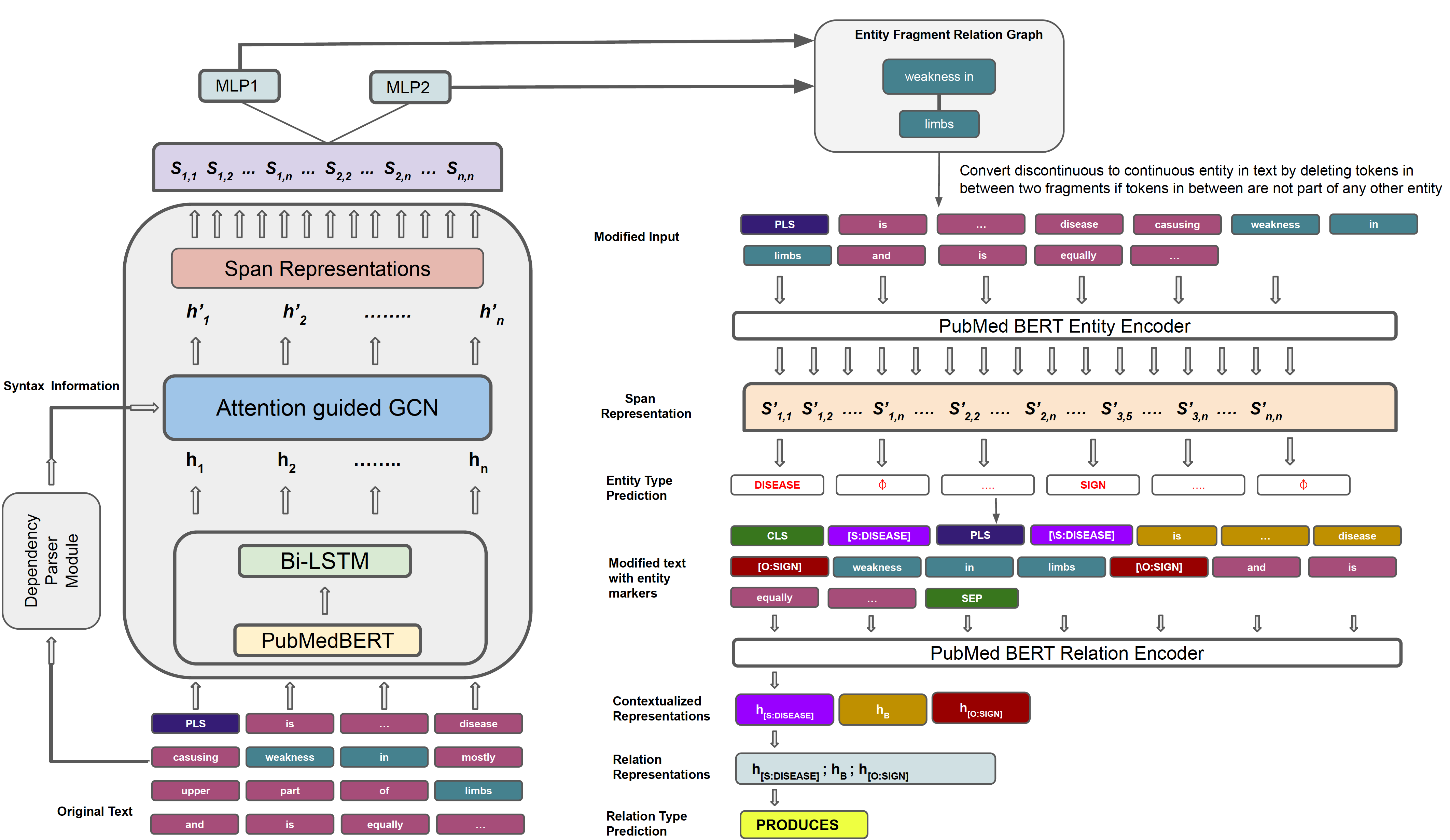}
\caption{\label{fig: pipeline} Pipeline approach using SODNER and PURE models for end-to-end relation extraction.}
\end{figure*}

One weakness of PURE is that it does not handle discontinuous entities in its NER component while it easily handles flat and nested entities. So we needed to adapt the PURE approach to the \textbf{\texttt{RareDis}} setting. Since PURE is pipeline-based, we could simply use a different NER model for identifying discontinuous entities and retain the PURE model to spot flat and nested entities. Hence, we use a specialized model that was exclusively developed for handling discontinuous entities called SODNER~\cite{li2021span}, which is also a span-based NER model that models discontinuous NER task as a classification problem to predict whether entity fragments with gaps ought to be linked to form a new entity. To do this, SODNER uses dependency parses of the input document to guide a graph convolutional neural (GCN) network that obtains enhanced contextual embeddings to link disparate fragments and form discontinuous entities. Figure~\ref{fig: pipeline} shows the schematic of the pipeline we use. It starts on the left with the SODNER model identifying discontinuous entities.

 Even if SODNER successfully identifies discontinuous entities, PURE's relation extraction model cannot handle them. The PURE relation model puts exactly one start and one end entity marker token around each candidate subject (or object) entity span. This modified input is passed through the contextual language model (such as PubMedBERT) and the marker token embeddings are used to predict the relation type. This is reflected by the purple [S:Disease] and [$\backslash$S:Disease] tokens on the right side of Figure~\ref{fig: pipeline}. But SODNER outputs multiple fragments for discontinuous entities. Rather than change the PURE relation model architecture, we use the discontinuous entity fragments and straightforward rules to convert the input sentence to a modified one where the discontinuous entities are rendered in a continuous format.  For instance, consider the input, ``weakness in the muscles of the arms and legs,'' which contains two entities: one flat entity, ``\textbf{weakness in the muscles of the arms} and legs'' and one discontinuous entity, ``\textbf{weakness in the muscles of the} arms and \textbf{legs}.'' Both entities have the gold entity type \textbf{Sign}. Our modified new input will read as: ``\textbf{weakness in the muscles of the arms} and \textbf{weakness in the muscles of the legs}''. This transformed sentence is now input through the PURE NER model and then through the PURE relation model. 

%Some limitations still exist in our pipeline. 
Neither the PURE NER model nor SODNER can handle cases where the same span has more than one entity type (e.g., a span being both a disease and a sign). This is a special case of overlapped entities where the overlap is exact, leading to the same span having two types. Since most relations involving such spans only use one of the entity types, this has not caused major issues in RE evaluation. 

\subsubsection{Sequence-to-Sequence: The Seq2Rel and T5 Model}
\label{sec-seq2rel}
The Seq2Rel model~\cite{giorgi2022sequencetosequence} model uses an encoder-decoder framework to process the input document and output relations akin to machine translation where the source language sentence is ingested into the encoder and the target language sentence is output by the decoder one token at a time. Here the target sequence is essentially a list of relations. Unlike the machine translation setting where the target is a natural language sequence where an order is inherent, relations do not have any order among them. Hence, during training an order is imposed on the relations in a document. Special tokens are also used to represent entity types. For example, the relation R2 in Figure~\ref{fig: RE_anno_change} indicates: (\textbf{Rare disease} ``Vitamin D Deficiency Rickets'', produces, \textbf{sign} ``bone disease''), where the entity types are in bold. This will be linearized in Seq2Rel as: Vitamin D Deficiency Rickets \texttt{@RareDisease@} bone disease \texttt{@Sign@} \texttt{@PRODUCES@}, where \texttt{@\textbf{ENTITY-TYPE}@} and \texttt{@\textbf{RELATION-TYPE}@} are special tokens indicating entity and relation types, respectively. The \texttt{@\textbf{ENTITY-TYPE}@} tokens are preceded by the actual entity spans in the input. If an input does not contain any relations, a special \texttt{@\textbf{NOREL}@} is coded as the output. 
The order imposed during training is simply the order in which the entities occur in the document. This is reflected in Figure~\ref{fig: RE_anno_change} where relations involving entities that occur earlier in the document are annotated before relations that involve entities that occur later. 
This left-to-right order is followed until all relations are output followed by a special end of sequence token \texttt{@\textbf{END}@}  signaling that all relations have been output. 
Besides this linearization schema, 
a ``copy mechanism''~\cite{zeng2018extracting} is applied to the decoder, restricting it to generate tokens only from the observed input sequence, unlike the full vocabulary of the target language in machine translation. This mechanism enables the decoder to output spans of the input text that correspond to entities, as well as special tokens representing relation labels that connect these entities. The Seq2Rel model~\cite{giorgi2022sequencetosequence} uses a PubMedBERT model as the encoder and a long short-term memory (LSTM) network as the decoder.

T5, developed by Google Research, challenges the conventional task-specific architectures by converting every NLP problem into a text-to-text input-output format. A key aspect of T5 is its baseline pre-training objective. For this, a large free text dataset known as the "Colossal Clean Crawled Corpus" was created and random spans of text are masked with the model tasked to predict these spans. Unlike masked language modeling in BERT models, each masked spans is replaced with only one sentinel token given a unique ID.   This approach helps the model learn a broad understanding of language and context. This baseline model is further trained on a suite a suite of NLP tasks (e.g., sentiment analysis, word sense disambiguation, and sentence similarity) in the text-to-text format. Another significant feature of T5 is its scalability, with versions ranging from small (60 million) to extremely large (11 billion), allowing it to be tailored to specific computational constraints and performance requirements.

Flan-T5 is an extension of T5 that is instruction fine-tuned on 1800 tasks. During this phase, the model is fine-tuned on a diverse range of tasks but with instructions provided in natural language. This training method enables Flan-T5 to understand and execute tasks based on straightforward instructions, making it more flexible and applicable to a wide range of real-world scenarios without requiring extensive task-specific data. It is fine-tuned both with and without exemplars (i.e., zero-shot and few-shot) and with and without chain-of-thought, enabling generalization across a range of evaluation scenarios. Please note that unlike Seq2Rel architecture, the outputs for T5 models variants are expected to following natural sentence structures, which are discussed in the next section as they are common to both T5 and GPT models.

\subsubsection{Generative Pre-trained Transformers: BioMedLM}
\label{sec:bioGPT_method}
Generative pre-trained transformers (GPTs) have captured the fascination of the general public and researchers alike, especially since the introduction of ChatGPT in December 2022. However, the in-context learning and few-shot capabilities have already surfaced in June 2020, when Open AI released GPT-3~\cite{brown2020language}. Building on the decoder component of the transformer architecture with the main objective of autoregressive left to right next token prediction task, they have excelled at text generation tasks (e.g., summarization). However, there is a growing interest in assessing their capabilities for language understanding tasks including relation extraction. 
BioGPT~\cite{Luo_2022} and BioMedLM~\cite{biomed-lm} have been pre-trained from scratch on biomedical abstracts from PubMed and full text articles from PubMed Central (from the corresponding subset from Pile~\cite{gao2020pile})  based on the GPT-2 model~\cite{radford2019language}. In this effort, we focus on BioMedLM, a 2.7B parameter  model,  comprised of 32 layers, a hidden size of 2560, and 20 attention heads. BioMedLM is an order of magnitude larger than BioGPT and nearly twice as large as BioGPT$_{large}$. It has been shown to be be superior to BioGPT models (including in our experiments for this paper where BioGPT underperforms by 10-15\% in F-score) and to our knowledge is the largest public GPT-style model for biomedicine. Hence, we only show BioMedLM results in this manuscript for the sake of clarify and simplicity. 
Unlike Seq2Rel whose sequence generation capabilities are highly constrained to terms observed in the input,  BioMedLM and BioGPT are purely generative, and supervised fine-tuning involves using appropriate prompts and output templates. Technically, we could simply use the linearization schemas introduced for Seq2Rel. However, these generative models generate natural language statements and not unnatural-looking templates. So our initial experiments using a Seq2Rel style output schemas have failed. So, we considered two types of schemas here:

\begin{itemize}
    \item \textbf{\texttt{rel-is}} template:  This output template is the same as that used by the original BioGPT paper for E2ERE: ``The relation between \textbf{subject-span} and \textbf{object-span} is \textbf{relationType.noun},'' where \textbf{relationType.noun} is the noun form of the predicate. With this template, as an example, the output for the gold relation (Wilm's tumor, is\_a, kidney cancer) is: ``The relationship between Wilm's tumor and kidney cancer is hyponym''. We can see here that we converted ``is a'' predicate to a noun representation ``hyponym'' in the template and a similar strategy was followed for all predicates.

    \item  \textbf{\texttt{natural-lang}}: We came up with different natural language templates tailored to each relation type in \textbf{\texttt{RareDis}}. They are fully specified in Table~\ref{tab:gpt-template}, each with a representative example.
\end{itemize}

\begin{table*}[ht]
\centering
\renewcommand{\arraystretch}{1.1}
\resizebox{.98\textwidth}{!}{
\begin{tabular}{lc}
\toprule
\multirow{2}{*}{\textbf{Relation type}} &  \textbf{Natural language output template}  \\
& (An example for the template)   \\ \midrule
\multirow{2}{*}{produces} &  \multicolumn{1}{c}{$ent_1Span$ is a {$ent_1Type$} that \textbf{produces} $ent_2Span$, as a {$ent_2Type$}}  \\
& (Asherman's syndrome is a rare disease that produces abdominal pain, as a symptom)   \\ \midrule
\multirow{2}{*}{anaphora} &  \multicolumn{1}{c}{The term $ent_2Span$ is an \textbf{anaphor} that refers back to the entity of the {$ent_1Type$} $ent_1Span$}  \\ 
& (The term ``it'' is an anaphor that refers back to the entity of the disease encephalitis)  \\ \midrule
\multirow{2}{*}{is\_synon} &  \multicolumn{1}{c}{The {$ent_1Type$} $ent_1Span$ and the {$ent_2Type$} $ent_2Span$ are \textbf{synonyms}}  \\
& (The disease diastrophic dysplasia and the rare disease disastrophic dwarfism are synonyms)  \\ \midrule
\multirow{2}{*}{is\_acron} &  \multicolumn{1}{c}{The \textbf{acronym} $ent_1Span$ stands for $ent_2Span$, a {$ent_2Type$}}  \\
& (The acronym LQTS stands for long QT syndrome, a rare disease)  \\ \midrule
\multirow{2}{*}{increases\_risk\_of} &  \multicolumn{1}{c}{The presence of the $ent_1Type$ $ent_1Span$ \textbf{increases the risk of} developing the {$ent_2Type$} of $ent_2Span$}  \\
& (The presence of the disease neutropenia increases the risk of developing the disease infections)  \\ \midrule
\multirow{2}{*}{is\_a} &  \multicolumn{1}{c}{The $ent_1Type$ $ent_1Span$ \textbf{is a} type of $ent_2Span$, a {$ent_2Type$}}  \\
& (The rare skin disease Bowen disease is a type of skin disorder, a disease)  \\ 

\bottomrule
\end{tabular}}
\caption{\label{tab:gpt-template} Natural language  templates used to encode \textbf{\texttt{RareDis}} relations as BioMedLM outputs.}
\end{table*}

\subsection{Training objectives and evaluation metrics}
For the SODNER+PURE pipeline model, the training objective is the well-known cross entropy function for both NER and RE components. Seq2Rel  and BioMedLM, however, produce sequences (based on the schemas and templates selected) that need to be interpreted back into the triple format (which we accomplish using regular expressions). Since their outputs are sequences, the training objective is the well-known auto-regressive language model objective based on predicting the next token given previously predicted tokens. The loss function is the average cross-entropy per target word (more details in Chapter 9.7 of Jurafsky and Martin~\cite{slpbook3}).

For evaluation, we note that \textbf{\texttt{RareDis}} annotations are at the span level and hence the same exact relation connecting the same entities can occur multiple times if it is discussed several times in the document. However, Seq2Rel and  BioMedLM  do not keep track of the number of times a relation occurs as they are generative and do not operate on spans; but the pipeline models output all connections as they operate at the span level. To ensure fair evaluation, if the same relation occurs multiple times within an instance, it is collapsed into a single occurrence. This is natural and harmless because there is no loss of information if duplicate relations are ignored. Since Seq2Rel and BioMedLM  produce sequences, we use regular expressions on top of the output templates and schemas to produce the triples we need. The evaluation metrics are precision, recall, and F1-score, which are standard in RE. For a relation to be counted as correctly predicted, the subject and object entity types, their spans, and the relation type all need to exactly match the ground truth relation. 

\section{RESULTS AND DISCUSSION}

Experiments for the pipeline approach were performed on our in-house cluster of 32GB GPU. All experiments for Seq2Rel were performed on Google Colab Pro+ using an Nvidia a100-sxm4-40gb GPU with access to high RAM. In Seq2Rel, we use AllenNLP, an open-source NLP library developed by the Allen Institute for Artificial Intelligence (AI2). Fairseq, a sequence modeling toolkit, is used for training custom models for text generation tasks for BioGPT on Google Colab Pro. We used \href{https://lambdalabs.com/}{Lambda Labs} to fine-tune BioMedLM on a single H100 80GB GPU. 

Next, we describe model configurations and hyperparameters. Our settings for learning rate, number of epochs, and other hyperparameters are determined based on experiments on the validation dataset.
\begin{itemize}
    \item Pipeline (SODNER+PURE): We used a batch size of 8, a learning rate of 1e-3, and 100 epochs to train the SODNER model for discontinuous entities with a PubMedBERT$_{base}$ encoder. For the PURE NER model, we used PubMedBERT$_{base}$ and trained for 100 epochs, with a learning rate of 1e-4 and a batch size of 8. We also experimented with PubMedBERT$_{large}$ with the same settings. For the PURE relation model,  we used both PubMedBERT$_{base}$ and PubMedBERT$_{large}$ as encoders with a learning rate of 1e-5  and trained for 25 epochs with the training batch size of 8. 
    
    \item Seq2Rel: Training was conducted for 150 epochs, with a learning rate of 2e-5 for the encoder (PubMedBERT$_{base}$ or PubMedBERT$_{large}$) and 1.21e-4 for the decoder (LSTM) with a batch size of 2 and a beam size of 3 (for the decoder).

    \item   BioMedLM: %We fine-tuned BioGPT and BioGPT$_{large}$. 
    Despite supervised fine-tuning, it is not uncommon for GPT models to output strings that were not part of the input. We observed that nearly 3\%-7\% of entities output by BioMedLM did not exactly match ground truth spans. Since we require an exact match for a prediction to be correct, we appended explicit natural language instructions to the input, directing the model to generate tokens from the input text: ``From the given abstract, find all the entities and relations among them. Do not generate any token outside the abstract.''
    We used a batch size of 1 with gradient\_accumulation\_steps of 16, a learning rate of 1e-5, and 30 epochs for BioMedLM.

    \item   T5: Using the same output templates   used for BioMedLM, we trained T5-3B, Flan-T5-Large (770M), and Flan-T5-XL (3B). For T5-3B, we used a batch size of 1 with gradient\_accumulation\_steps set to 16, lr = 3e-4, 100 epochs, and generation beam size of 4. For Flan-T5, we used a batch size of 2 with gradient\_accumulation\_steps set to 16, and the rest of the hyperparameters same as T5-3B. For Flan-T5-XL, we used a batch size of 1 with gradient\_accumulation\_steps set to 16, lr = 3e-4, 100 epochs, and generation beam size of 4 with DeepSpeed for CPU offloading of the parameters.

\end{itemize}

We also needed some post-processing tricks to handle the idiosyncrasies of the three different models. As we discussed earlier in Section~\ref{sec-pipeline}, for the pipeline models, since discontinuous entities are not handled natively by the PURE relation model, we had to transform the inputs to render the discontinuous entities in a flat fashion before passing them on to the PURE model. For the Seq2Rel model, due to the WordPiece tokenization in BERT models, the output sometimes contains extra spaces around hyphens and brackets.  To align such output strings with the input text, as a post-processing step, we removed these additional spaces, specifically around hyphens, curved brackets, and forward slashes. For the rel-is template, T5  and its variant were predicting the synonym relation with the string \textbf{``synonyms''}; so as a part of the post-processing, we replaced with with ``synonym.''

The main results of the comparison using different models are presented in Table~\ref{results}. For   BioMedLM and T5 models, the `copyInstruct' column in the table indicates the additional input prompt discussed earlier in this section where models are directed to only generate tokens observed in the input. We observe that the SODNER+PURE pipeline (with PubMedBERT$_{base}$ encoder) produces the best F1-score  of 52.2, which is 5 points more than the best-performing Seq2Rel model with the PubMedBERT$_{large}$ encoder (47.15 F1),  5.2 points better than the best-performing model from T5 family (Flan-T5-large), and 13 points more than best performing BioMedLM model (38.9 F1).
The pipeline's performance does not increase when using the PubMedBERT$_{large}$ model.  
For Seq2Rel, using PubMedBERT$_{large}$ outperforms a model with PubMedBERT$_{base}$ (44.53 F1) by 2.5 points, with an increase in both precision and recall. Potentially, the increased model capacity of  PubMedBERT$_{large}$ enables it to capture more complex and subtle relationships between medical terms and concepts. However, it is not clear why similar gains were not observed with PubMedBERT$_{large}$ in the pipeline. 

\begin{table*}[ht]
\renewcommand{\arraystretch}{1.2}
\centering
%\resizebox{0.7\linewidth}{!}{%
\begin{tabular}{c c c  c c c}
\toprule
\multirow{2}{*}{ Method} & \multirow{2}{*}{ Configuration} & \multirow{2}{*}{ copyInstruct} &  \multicolumn{3}{c}{Score} \\
\cmidrule(r){4-6}
& & &   P & R & F \\
\cmidrule{1-6}
 SODNER + PURE & PubMedBERT$_{base}$ & NA &  55.99 & \textbf{48.89} & \textbf{52.20} \\
 SODNER + PURE & PubMedBERT$_{large}$ & NA & \textbf{56.20} & 48.52 & 52.08 \\
\midrule
\multirow{2}{*}{Seq2Rel} & PubMedBERT$_{base}$ & NA  &  47.60 & 40.90 & 44.53 \\
& PubMedBERT$_{large}$ &  NA &  51.46 & 43.51 & 47.15 \\
\midrule
\multirow{4}{*}{Flan-T5-Large} & \textbf{\texttt{rel-is}} & yes &  46.52 & 46.58 & 46.55 \\
& \textbf{\texttt{rel-is}} & no &  48.63 & 45.54 & 47.04  \\
& \textbf{\texttt{natural-lang}} & yes & 43.83 & 42.82 & 43.32 \\
& \textbf{\texttt{natural-lang}} & no &  40.07 & 40.17 & 40.12  \\
\midrule
\multirow{4}{*}{T5-3B} & \textbf{\texttt{rel-is}} & yes &  41.13 & 39.36 & 40.22 \\
& \textbf{\texttt{rel-is}} & no &  45.72 & 41.50 & 43.51  \\
& \textbf{\texttt{natural-lang}} & yes & 44.25 & 40.71 & 42.40 \\
& \textbf{\texttt{natural-lang}} & no &  37.80 & 41.21 & 39.43  \\
\midrule
\multirow{4}{*}{Flan-T5-XL} & \textbf{\texttt{rel-is}} & yes &  45.00 & 40.82 & 42.82 \\
& \textbf{\texttt{rel-is}} & no & 44.16 & 38.10 & 40.91  \\
& \textbf{\texttt{natural-lang}} & yes & 44.68 & 42.87 & 43.76 \\
& \textbf{\texttt{natural-lang}} & no &  42.05 & 40.87 & 41.45  \\
\midrule
\multirow{4}{*}{ BioMedLM} & \textbf{\texttt{rel-is}} & yes &  40.19 & 29.68 & 34.14 \\
& \textbf{\texttt{rel-is}} & no &  42.14 & 36.1 & 38.89  \\
& \textbf{\texttt{natural-lang}} & yes & 38.64 & 32.81 & 35.49 \\
& \textbf{\texttt{natural-lang}} & no &  44.22 & 33.76 & 38.29  \\
\bottomrule
\end{tabular}%
%}
\caption{Performances of different models under different settings on the \textbf{\texttt{RareDis} dataset}. \label{results}}
\end{table*}

 The best performance for BioMedLM is an F1 score of 38.89 using the \textbf{\texttt{rel-is}} template for prompting the model when copy instructions were not provided. 
When copy instructions are not provided, \textbf{\texttt{rel-is}} does slightly better (<1\% F1) and when copy instructions are not provided, \textbf{\texttt{natural-lang}} does better job (1.35 of points gain)
So looks like there is no advantage to using copy instructions. (However, when using the smaller \textbf{BioGPT} models, the natural language prompting seemed to perform slightly better than the \textbf{\texttt{rel-is}} template.)
Note that, BioMedLM's best performance is still $\approx 6$ points lower than then Seq2Rel's best score and 11 points lower than the pipeline score.

Note that BioMedLM is over eight times larger  than our best-performing pipeline model (considering it has three encoders based on the encoder   PubMedBERT$_{base}$, which has 110M parameters). 
However, its low performance compared to the pipeline is not surprising because GPT models are autoregressive and do not benefit from language understanding arising from the bidirectional masked language modeling objective used in BERT models. 
Although the original BioMedLM~\cite{biomed-lm} effort did not perform RE, it reports SOTA scores on biomedical Q\&A tasks. The smaller BioGPT models were shown to do better than BERT models for E2ERE too. 
Hence we repurposed them for this RE task and as the largest publicly available GPT-based model, BioMedLM outperformed BioGPT models~\cite{Luo_2022} by 10--15\% in F1 score and we do not see these as worthy of reporting in this manuscript.

The best-performing model from the T5 family is Flan-T5-large with an F1 score of 47 using the \textbf{\texttt{rel-is}} template for prompting the model when copy instructions were not provided, which is the same configuration that worked best for BioMedLM. It is surprising to see that even though Flan-T5-Large (780M) is much smaller than T5-3B and Flan-T5-XL (3B), it outperforms the other two in every setting, except Flan-T5-XL with the natural-lang template. On comparing the same size T5 models (T5-3B and Flan-T5-XL),  Flan-T5-XL performs better in most settings. 
%On the other hand, Flan-T5-XL outperforms T5-3B for the settings where copy instruction was provided for both templates. 
%This result is surprising as first, Flan-T5-XL is instruction finetuned which should have outperformed T5-3B in every perspective and second, it did not beat its smaller finetuned version (Flan-T5-XL) in rel-is template settings.
We believe much larger models (GPT-3, GPT-3.5, GPT-4) ought to be used to fully leverage the power of generative LMs. Furthermore, some recent results also show that using GPT-style models to generate additional training examples to augment the training data may be a more effective way of using them, rather than fine-tuning them for RE tasks.

\begin{table*}[htbp]
\renewcommand{\arraystretch}{1}
\centering
\resizebox{\linewidth}{!}{%
\begin{tabular}{l r r r r r r r r r r r r r r r}
\toprule
\multicolumn{1}{l}{\multirow{2}{*}{Relation type}} & \multicolumn{3}{c}{SODNER+PURE} & \multicolumn{3}{c}{Seq2Rel}  & \multicolumn{3}{c}{BioMedLM} & \multicolumn{3}{c}{Flan-T5-large}  \\ 
\cmidrule{2-13} 
\multicolumn{1}{c}{} & \multicolumn{1}{c}{P} & \multicolumn{1}{c}{R} & \multicolumn{1}{c}{F} & \multicolumn{1}{c}{P}   & \multicolumn{1}{c}{R}   & \multicolumn{1}{c}{F}  & \multicolumn{1}{c}{P}   & \multicolumn{1}{c}{R} & \multicolumn{1}{c}{F}   & \multicolumn{1}{c}{P} & \multicolumn{1}{c}{R} & \multicolumn{1}{c}{F} \\ 
\midrule
anaphora & {70.40} & {69.84} & {\textbf{70.11}} & {64.60} & {58.00} & 61.08 & 61.26 & 53.96 & 57.38 & 62.99 & 63.49 & 63.24\\ 
is\_a & {62.67} & {55.29} & \textbf{58.75} & {58.67} & {51.76} & 55.00 & 52.77 & 44.70 & 48.40 & 61.84 & 55.29 & 58.38\\ 
is\_acron & {70.37} & {57.58} & \textbf{63.33} & {50.00} & {42.00} & 45.65 &  55.17 & 48.48 & 51.61 & 59.25 & 48.48 & 53.33\\ 
produces & {50.21} & {45.09} & \textbf{47.51} & {47.48} & {41.13} & 44.00 & 37.20 & 32.82 & 34.87 & 43.05 & 43.45 & 43.24 \\ 
is\_synon & {75.00} & {18.75} & \textbf{30.00} & {100.00} & {12.50} & 22.23 &  {0.00} & {0.00} & 0.00 & 0.00 & 0.00 & 0.00 \\ 
increases\_risk\_of & {50.00} & {4.55} & 8.33 & {11.80} & {9.52} & \textbf{{10.52}} & {0.00} & {0.00} & 0.00 & 0.00 & 0.00 & 0.00 \\
\bottomrule
\end{tabular}%
}
\caption{Scores for each relation type of best-performing models in the group.\label{results_rel_types}}
\end{table*}

We also wanted to examine scores per relation type in our models to see if there are any predicates for which we are underperforming more than expected. From 
Table~\ref{results_rel_types}, we notice that recall is less than 5\% for 
increases\_risk\_of relation type. This is quite awful but not surprising given the prevalence of such relations is very small in the dataset (from Table~\ref{tab:ent_rel_types}). But what is very unusual is the F1 of the `produces' relation being less than 50, when it constitutes over 60\% of all relations in the dataset (from Table~\ref{tab:ent_rel_types}). Upon deeper investigation, we found that generally longer object entities lead to NER errors. We checked this more concretely by examining the errors (for `produces') and found out that we missed 43\% of the object spans for the best-performing pipeline method. Thus, a large portion of performance loss is simply due to the model not being able to predict the object entity span correctly; especially for long object entities, even missing a single token can lead to RE errors. 

Thus, the overall performance pattern observed for the  \textbf{\texttt{RareDis}} dataset is Pipeline $>$ Seq2Rel $>$ Flan-T5-Large $>$ Flan-T5-XL $>$ T5-3B $>$ BioMedLM. We wanted to verify this with at least one other dataset. Considering our prior experiences with the chemical-protein interaction extraction task~\cite{ai2023end}, we repeated our E2ERE experiments using the BioCreative Shared Task VI dataset and the results showed the same performance pattern with pipeline leading to a 69 F1 score, followed by Seq2Rel with 49, and BioMedLM with 37 points.

\section{Error Analysis} 
Before we proceed, we note that many RE errors appear to arise from NER errors. This can lead to a snowball effect of errors in the RE phase. Consider a single entity participating in $n$ gold relations. If it is predicted incorrectly as a partial match, it may potentially lead to $2n$ relation errors because it can give rise to $n$ false positives (FPs) (because the relation is predicted with the wrong span) and $n$ false negatives (FNs) (because the gold relation with the right span is missed). Thus, even a small proportion of NER errors can lead to a high loss in RE performance. In this section, we discuss a few error categories that we observed commonly across models. 
\begin{itemize}
\item \textbf{Partial matches:} When multi-word entities are involved, the relation error is often due to the model predicting a partial match (a substring or superstring of a gold span) and this was frequent in our effort. Consider the snippet \textit{``Kienbock disease changes may produce pain...The range of motion may become restricted''}. Here Kienbock disease is the subject of a \textit{produces} relation with the gold object span: ``the range of motion may become restricted''. However, the Seq2Rel model predicted ``range of motion restricted'' as the object span, leading to both an FP and FN. But common sense tells us that the model prediction is also correct (and potentially even better) because it removed the unnecessary ``may become'' substring. In a different example, when the relation involved the gold span ``neurological disorder,'' the model predicted a superstring ``progressive neurological disorder'' from the full context: ``Subacute sclerosing panencephalitis (SSPE) is a progressive neurological disorder.'' 

\item \textbf{Entity type mismatch: } Because our evaluation is strict, predicting the entity spans and relation type correctly, but missing a single entity type can invalidate the whole relation leading to both an FP and an FN. 
The models are often confused between closely related entity types. \textbf{Rare disease} and \textbf{skin rare disease} were often confused along with the pair \textbf{sign} and \textbf{symptom}. 

\item \textbf{Issues with discontinuous entities:} Discontinuous entities are particularly tricky and have led to several errors, even if the prediction is not incorrect, because the model was unable to split an entity conjunction into constituent entities. Consider the snippet: \textit{``affected infants may exhibit abnormally long, thin fingers and toes and/or deformed (dysplastic) or absent nails at birth.''} Instead of generating relations with the two gold entities ``abnormally long, thin fingers'' and ``abnormally long, thin toes'', the model simply created one relation with ``long, thin fingers and toes.''

\item \textbf{ BioMedLM generations not in the input:} In several cases we noticed spans that were not in the input but were nevertheless closely linked with the gold entity span's meaning. For example, for the gold span ``muscle twitching'', BioMedLM predicted ``muscle weakness''. It also tried to form meaningful noun phrases that capture the meaning of longer gold spans. For instance, for the gold span ``ability to speak impaired'', it predicted ``difficulty in speaking''. For the gold span, ``progressive weakness of the muscles of the legs'' it outputs ``paralysis of the legs''. All these lead to both FPs and FNs, unfortunately.

\item \textbf{Errors due to potential annotation issues:} In document-level RE settings, it is not uncommon for annotators to miss certain relations. But when these are predicted by a model, they would be considered FPs. Consider the context: \textit{``The symptoms of infectious arthritis depend upon which agent has caused the infection but symptoms often include fever, chills, general weakness, and headaches.''} Our model predicted that ``infectious arthritis'' \textit{produces} ``fever''. However, the gold predictions for this did not have this and instead had the relation ``the infection'' (anaphor) \textit{produces} ``fever''. While the gold relation is correct, we believe what our model extracted is more meaningful. However, since we missed the anaphor-involved relation, it led to an FN and an FP. 

\end{itemize}

\section{Conclusion}
In this paper, we explored four state of the art representative models for E2ERE from three competing paradigms: pipelines (SODNER + PURE), sequence-to-sequence models (Seq2Rel, T5), and generative LMs (BioMedLM). Our evaluations used a complex dataset (\textbf{\texttt{RareDis}}) involving discontinuous, nested, and overlapping entities. Even with the advances in Seq2Seq models and generative transformers, a custom-built pipeline still seems to be the best option based on our experiments in this paper.  The performance gap between Seq2Rel and the pipeline is not as high as that between BioMedLM and pipeline. As such there could be other datasets where Seq2Rel matches the pipeline methods especially for simpler NER scenarios without discontinuous entities. We still would not want readers to conclude that more advanced models are not suitable for this task and not to take away from the few-shot abilities of GPT models. Also, the generative aspects of GPT models may not be suitable for the type of strict evaluation imposed here where an exact match with gold spans is required. In the future, this may be mitigated by using vector similarity or edit-distance metrics to map such phrases to the closest matches of the input.
Using inference-only proprietary large models such as GPT-4~\cite{bubeck2023sparks} to generate paraphrases for training instances to create larger augmented training datasets could also be helpful.  However, in the end, a small $\approx$ 200M parameter pipeline model that can run on consumer desktops may be preferable for several use-cases even in the current era of excitement over generative transformers.

\section*{Acknowledgment}
This work is supported by the NIH National Library of Medicine through grant R01LM013240.  The content is solely the responsibility
of the authors and does not necessarily represent the official views of the NIH.

\bibliographystyle{vancouver}
\bibliography{./sample} 
\end{document}